\documentclass[conference]{IEEEtran}
\IEEEoverridecommandlockouts
\usepackage{cite}
\usepackage{amsmath,amssymb,amsfonts}
\usepackage{algorithmic}
\usepackage{graphicx}
\usepackage{textcomp}
\usepackage{xcolor}
\def\BibTeX{{\rm B\kern-.05em{\sc i\kern-.025em b}\kern-.08em
    T\kern-.1667em\lower.7ex\hbox{E}\kern-.125emX}}
\begin{document}

\title{Segmentation of Ischemic Stroke Lesions using Transfer Learning on Multi-sequence MRI}

\author{\IEEEauthorblockN{Rishiraj Paul Chowdhury}
\IEEEauthorblockA{\textit{Department of Computer Science} \\
\textit{University of Colorado Boulder}\\
Boulder, United States \\
rishiraj.paulchowdhury@colorado.edu}
\and
\IEEEauthorblockN{Tanjim Rahman}
\IEEEauthorblockA{\textit{Department of Eletrical and Computer Engineering} \\
\textit{Texas Tech University}\\
Lubbock, United States \\
tanjrahm@ttu.edu}
}

\maketitle

\begin{abstract}
The accurate understanding of ischemic stroke lesions is critical for efficient therapy and prognosis of stroke patients. Magnetic resonance imaging (MRI) is sensitive to acute ischemic stroke and is a common diagnostic method for stroke. However, manual lesion segmentation performed by experts is tedious, time-consuming, and prone to observer inconsistency. Automatic medical image analysis methods have been proposed to overcome this challenge. However, previous approaches have relied on hand-crafted features that may not capture the irregular and physiologically complex shapes of ischemic stroke lesions. In this study, we present a novel framework for quickly and automatically segmenting ischemic stroke lesions on various MRI sequences, including T1-weighted, T2-weighted, DWI, and FLAIR. The proposed methodology is validated on the ISLES2015 Brain Stroke sequence dataset, where we trained our model using the Res-Unet architecture twice: first, with pre-existing weights, and then without, to explore the benefits of transfer learning. Evaluation metrics, including the Dice score and sensitivity, were computed across 3D volumes. Finally, a Majority Voting Classifier was integrated to amalgamate the outcomes from each axis, resulting in a comprehensive segmentation method. Our efforts culminated in achieving a Dice score of 80.5\% and an accuracy of 74.03\%, showcasing the efficacy of our segmentation approach.
\end{abstract}


\section{Introduction}
Stroke is a devastating global health concern, with ischemic stroke accounting for the majority (80–85\%) of cases. Ischemic stroke is caused by the occlusion of an artery leading to the brain, which results in diminished blood supply and can cause irreversible tissue damage. Effective patient treatment and prognosis are critically dependent on prompt intervention within a narrow therapeutic window (typically 4.5 hours) to mitigate disability and fatality. The crux of diagnosis and treatment lies in accurately determining the location and extent of the stroke lesion, a task known as lesion segmentation.

Magnetic Resonance Imaging (MRI) and Computed Tomography (CT) are the primary diagnostic modalities for ischemic stroke, where MRIs, specifically through the use of multi-sequence data (e.g., T1-weighted, T2-weighted, DWI, FLAIR), exhibit superior sensitivity in detecting ischemic stroke lesions. An accurate identification and volumetric analysis of lesions plays a crucial role in the treatment of ischemic stroke. However, the current standard of manual segmentation by neurologists is laborious, time-intensive, and the precision can be compromised by factors such as tissue structure complexity and ambiguous lesion characteristics. Thus, developing an automated system is imperative yet challenging, as previous approaches have often relied on traditional image processing or shallow models\cite{b3}\cite{b4} that fail to capture the irregular, physiologically complex lesion shapes effectively\cite{b10}\cite{b11}. Also, recent advancements in deep learning have enabled the automatic segmentation of ischemic stroke lesions by leveraging the deep architecture of convolutional neural networks (CNNs). They facilitate a fully automated process but face several challenges in this semantic segmentation task due to inherent limitations of medical data, including data scarcity,\cite{b8} class imbalance at the pixel level, variability in lesion location and size, and high similarity of lesions to imaging artifacts. Although the potential of Transfer Learning has been noted in the wider medical field,\cite{b6} the evaluation gap of TL benefits for this segmentation task necessitates the development of a robust, feature-learning architecture that can generalize effectively.

We thus propose a fully automated and novel framework that aims to solve the challenges faced in the accurate segmentation of ischemic stroke lesions from multi-sequence MRIs. Our framework is based on deep CNNs that utilize the Res-UNet architecture for pixel-wise semantic segmentation, which has been proven effective in medical imaging tasks by combining the U-Net's structure with residual connections. \cite{b7}\cite{b11} We focused on exploring transfer learning by training the model both with and without pre-existing weights. This is to assess the model efficacy in overcoming the limited availability of annotated medical imaging datasets. \cite{b5}\cite{b6} Also, to enhance 3-D MRI image consistency against different artifacts, we combine the segmentation outcomes from multi-imaging axes using Majority Voting Classifier (MVC). This fusion technique is designed to use multi-plane spatial information based on concepts used in other segmentation contexts. \cite{b3}\cite{b4} The primary novelty of our research lies in leveraging data from different perspectives to enhance the segmentation precision.

The rest of this paper is organized as follows: Section II details the methodology used and the architecture of the Res-UNet model. Section III describes the experimental setup, discusses the results obtained and compares the models trained with and without transfer learning. Finally, Section IV concludes the paper and outlines directions for future work.

\section{Methodology}

\subsection{Proposed Method}
\noindent Multiple Model Training: We have the 3D NIfTI format MRI sequence images (T1-weighted, T2-weighted, FLAIR, and DWI), where for each sequence we take 2D slices of them along the axial, sagittal, and frontal planes and feed the input to the 3 models for ModelX (x-plane), ModelY (y plane), ModelZ (z-plane), all of which are a Hybrid Res-UNet model pretrained on the ImageNet dataset. This uses the benefits of Transfer Learning for better feature extraction and Skip Connections to retain the spatial information. Later, the Decoder part up-samples the image, and the softmax layer at the end gives the array of pixel values localizing the lesion region and segments it along the three planes.\\
Prediction Phase: After fine-tuning each instance on the specific task of ischemic stroke lesion segmentation, each model makes its own predictions for lesion segmentation based on the input brain image planes (axial, sagittal, and coronal). All the slices across these three planes are stacked up to give the 3D segmented mask for X-plane, Y-plane, and Z-plane.\\
Voting Phase: Utilize the Majority Voting algorithm to combine the individual predictions from each model. For each voxel in the image, count the votes (i.e., predictions) from all models. The final prediction for that voxel would be the label with the majority of votes. The final output is to produce a single 3D segmentation mask representing the final decision about the lesion.
\begin{figure}[h]
    \centering
    \includegraphics[width=0.5\textwidth]{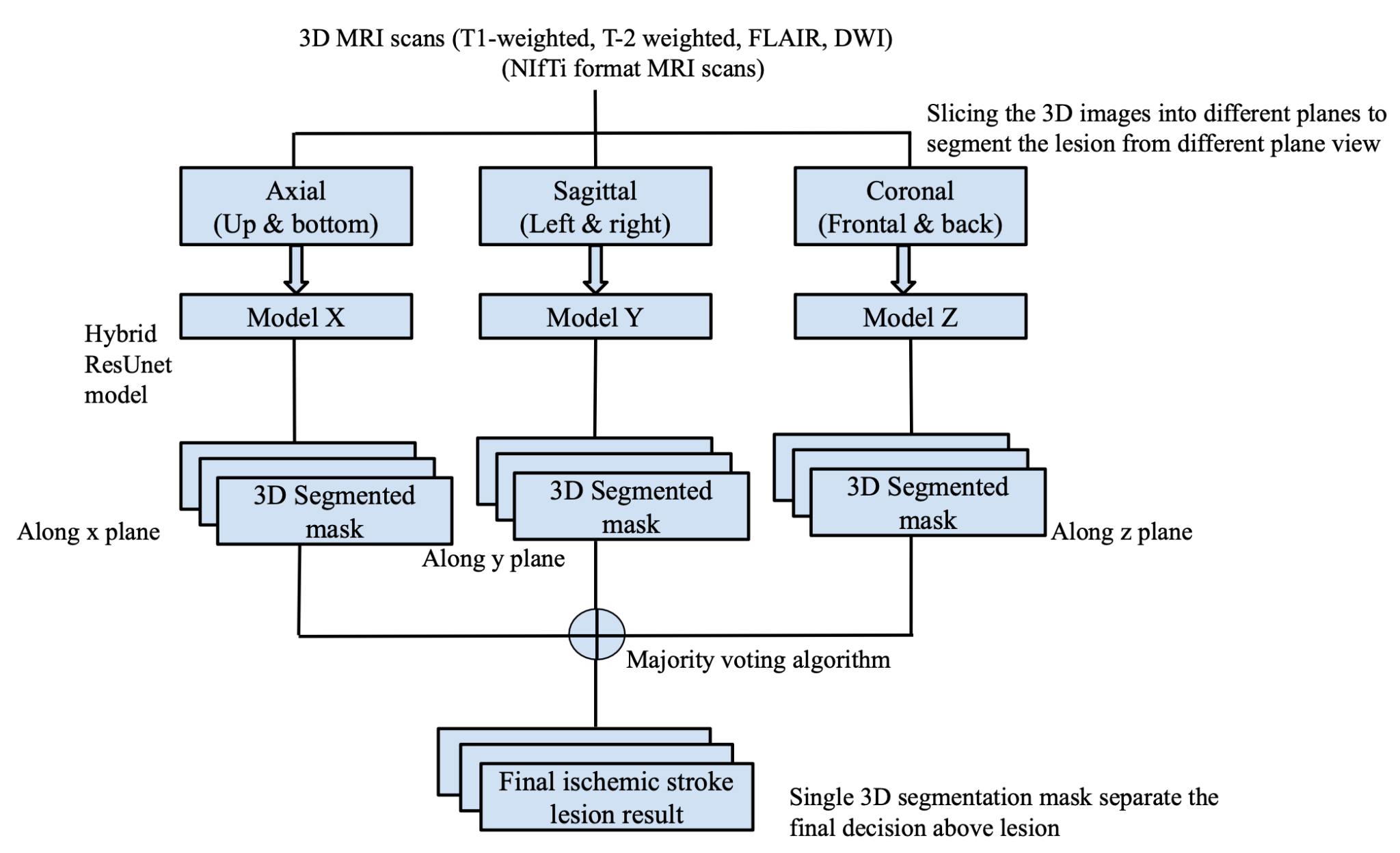}
    \caption{Workflow Diagram of our Proposed Segmentation  Method}
\end{figure}

This approach leverages the diversity of models generated from different initializations or hyperparameters and combines their predictions to potentially improve the accuracy of stroke lesion detection. It ensures that the final segmentation incorporates the complementary details from each plane while benefiting from the ensemble learning provided by the Majority Voting Classifier.

\subsection{Encoder-Decoder Structure}
U-Net: The U-Net architecture, crafted specifically for image segmentation duties, stands as a potent convolutional neural network (CNN) model. The UNet architecture is widely employed for 2D and 3D medical image segmentation, initially designed for microscopy cell segmentation and later extended to various medical imaging applications, particularly in brain tumors. It consists of two paths: an encoder for extracting relevant features through convolution layers and down-sampling, and a decoder for reconstructing probability segmentation maps.
\begin{figure}[h]
    \centering
    \includegraphics[width=0.5\textwidth]{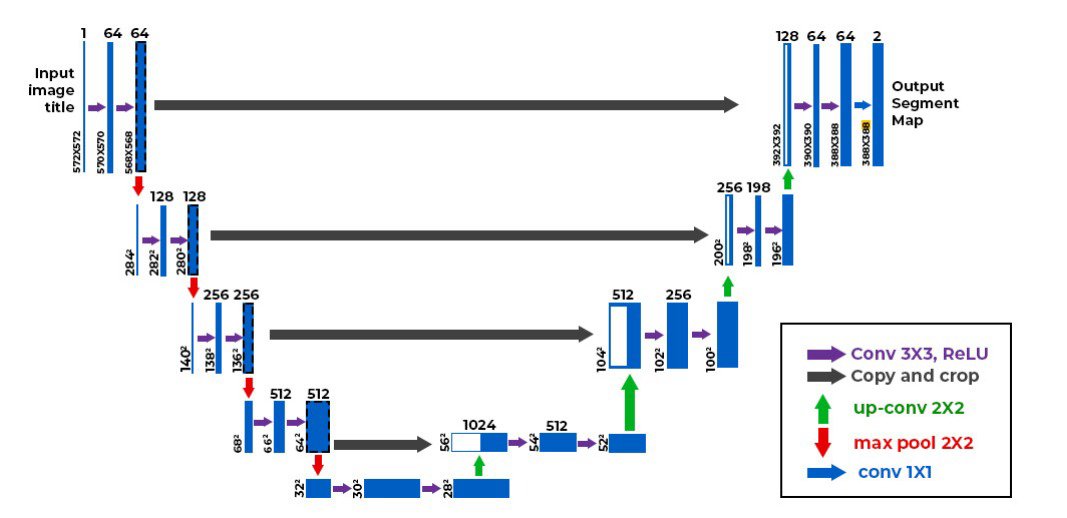}
    \caption{Architecture Diagram of a Baseline UNet Model}
\end{figure}\\
Encoder (Contracting Path):\\
• Processes the input image to extract high-level features.\\
• Uses multiple convolutional layers with activation functions (like ReLU) to learn patterns from the image.\\
• Employs pooling layers (like max pooling) to down-sample the image size, reducing spatial resolution but increasing the number of feature channels. This allows the network to capture broader contextual information.\\
Decoder (Expanding Path):\\
• Aims to recover spatial resolution while preserving the extracted features.\\
• Utilizes up-sampling layers (like transposed convolution) to increase the image size.\\
• Incorporates skip connections that directly concatenate feature maps from the corresponding encoder block at the same scale. This injects precise location information from the earlier stages, aiding in accurate segmentation of detailed structures.\\
• Employs additional lavers to refine the segmentation mask.

Skip Connections: These connections directly bridge the gap between the encoder and decoder at corresponding levels with the same spatial resolution. They act like information highways, allowing the decoder to access precise location details from the earlier stages of the encoder. This is crucial for accurate segmentation of fine-grained structures in the image, as the encoder might lose some spatial details during pooling. By incorporating skip connections, U-Net effectively combines high-level features from the encoder with the precise spatial information from the decoder, leading to superior segmentation results.

\subsection{ResNet and Hybrid ResNet Architecture}
ResNet, an abbreviation for Residual Network, is a robust deep learning structure renowned for its efficacy in effectively training extremely deep neural networks. It excels at learning hierarchical features from images and can be very deep due to its residual connections, which helps alleviate the vanishing gradient problem. Widely used and pre-trained models available for transfer learning. We can use a pre-trained ResNet as the encoder (feature extractor) in the U-Net model. This leverages the power of ResNet's feature learning while maintaining U-Net's segmentation capabilities.

The ResNet50 backbone is pretrained on the ImageNet dataset using transfer learning and fine-tuning tools. Our approach combines the encoder-decoder structure of UNet with ResNet50's residual units. The encoder layer is frozen using fine-tuning and transfer learning with the pretrained ResNet50 model on ImageNet, reducing the number of trainable parameters.
\begin{figure}[h]
    \centering
    \includegraphics[width=0.45\textwidth]{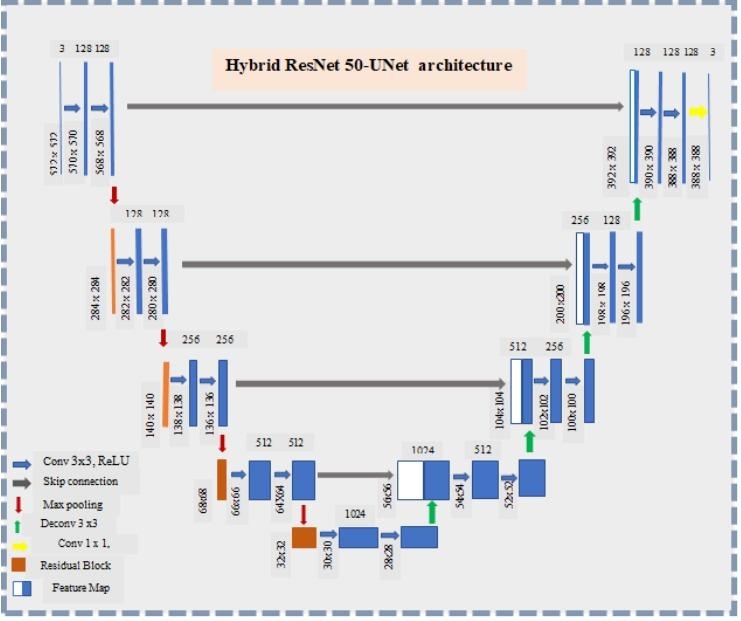}
    \caption{Architecture Diagram of Hybrid ResNet 50-UNet}
\end{figure}\\
Training with pre trained weights: When utilizing pre-trained weights for ResNet, the encoder benefits from learned features extracted from large-scale datasets, such as ImageNet. By leveraging these pre-trained weights, the encoder initializes with meaningful representations, enabling the model to capture intricate patterns and structures in the input images effectively. During training, the weights of the encoder remain frozen, allowing the model to focus on learning the decoder's parameters. This strategy accelerates convergence and improves generalization, particularly when the target dataset shares similar characteristics with the pre training data. Additionally, fine-tuning the entire network with a lower learning rate can further refine the model's performance, adapting it to the specific nuances of the segmentation task.\\
Training without weights: Training the ResNet encoder and UNet decoder from scratch without pre-trained weights requires the model to learn feature representations solely from the target dataset. While this approach may be computationally demanding and require more data for effective training, it offers the advantage of tailoring the model to the specific characteristics of the segmentation task at hand. Without relying on pre-trained weights, the network learns features directly relevant to the target domain, potentially leading to better alignment with the nuances of the segmentation problem. However, training from scratch typically necessitates a more extensive dataset and longer training times to achieve comparable performance to pre-trained models. Despite the initial challenges, training without pre-trained weights provides flexibility and adaptability, making it suitable for scenarios where pre-trained models may not be readily available or applicable.

\subsection{Majority Voting Classifier}
Apply a voting strategy such as majority voting to combine the individual predictions. For each pixel in the final segmentation mask from all the three models:

\noindent
• Count the number of models that predicted that pixel as part of the lesion.\\
• Assign the final label (lesion or background) based on the majority vote across all models. This creates a single, more robust segmentation mask representing the combined knowledge of the ensemble.
\begin{figure}[h]
    \centering
    \includegraphics[width=0.5\textwidth]{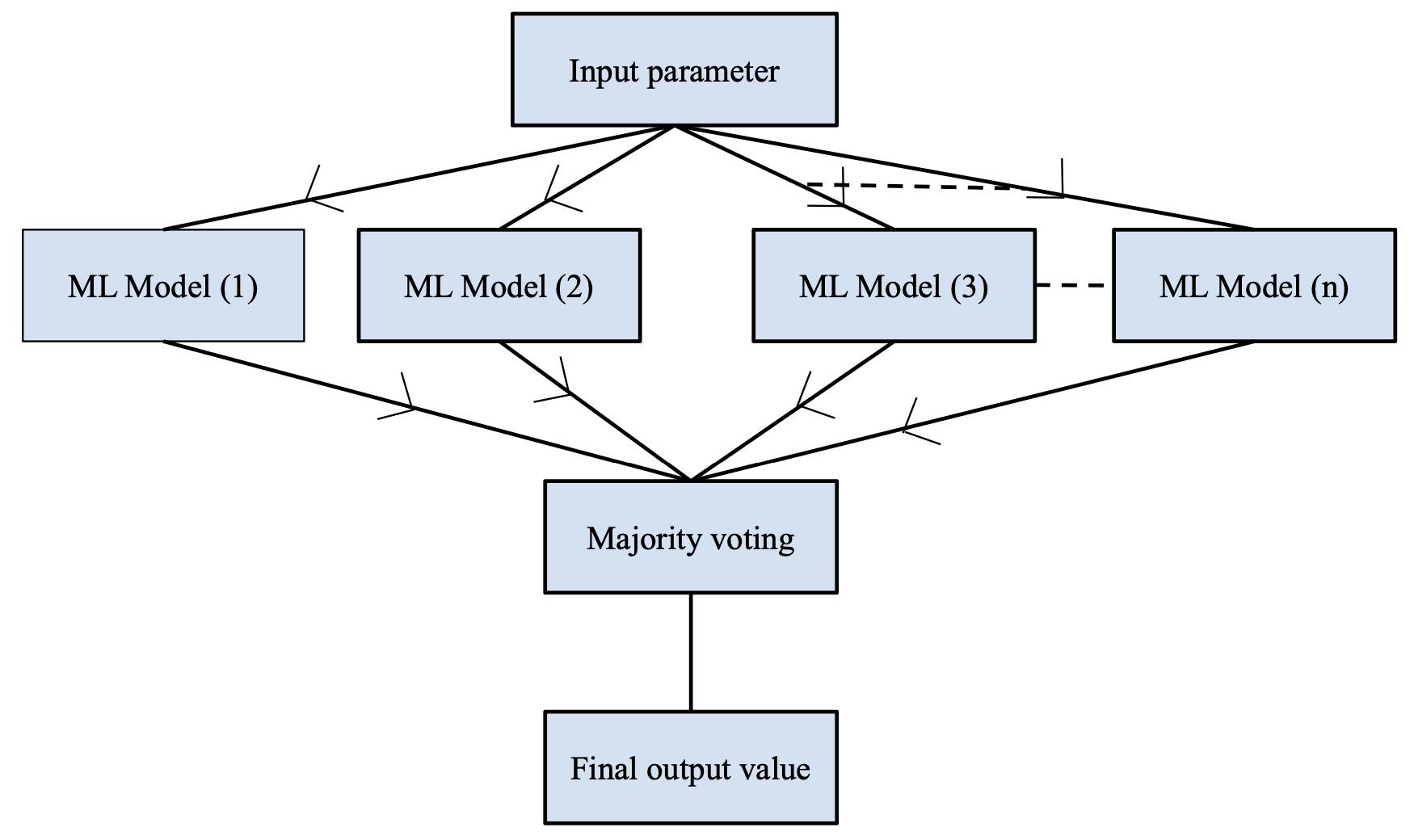}
    \caption{Flowchart Diagram of Majority Voting Classifier Algorithm}
\end{figure}
\\
There are various advantages to this approach.\\
• Improved Accuracy: By combining information from multiple planes, we can potentially capture more details about the lesion's shape and boundaries.\\
• Reduced Bias: Each plane might have limitations in visualizing specific lesion regions. Combining information reduces the dependence on any single plane for accurate segmentation.\\
• Robustness: If one plane's prediction is inaccurate due to noise or artifacts, the information from other planes can help compensate and provide a more reliable segmentation result.

\subsection{Loss Function}
Class imbalance is a common problem in medical image analysis due to the small size of lesions relative to the overall volume. To address this issue, we utilized a hybrid loss function during network training that combines weighted cross-entropy and dice loss. The dice coefficient is a commonly used overlap metric that measures the similarity between two sets of data, such as the predicted segmentation map and the ground truth.\\
The dice coefficient between two binary volumes can be written as:
\begin{align}
    DICE = \frac{2\sum^N_i{p_ig_i}}{\sum^N_i{p_i^2}+\sum^N_i{g_i^2}}
\end{align}
The network parameters were optimized to minimize the total loss, where the sums were taken over N voxels of the predicted binary segmentation volume $p_i$ in P and the ground truth binary volume $g_i$ in G.\\
The cross entropy loss is,
\begin{align}
    H(p,q) = -\sum_{x\ \epsilon\ classes}{p(x).log\ q(x)}
\end{align}
where p(x) = true probability distribution and q(x) = model’s predicted probability distribution.
Therefore, the Total Loss Function is,
\begin{align}
    Total \ loss = \lambda(cross \ entropy \ loss) + \gamma(dice \ loss)
\end{align}
where $\lambda$ and $\gamma$ are empirically assigned weights to individual losses. In this work, we set $\lambda = 0.50$ and $\gamma = 0.5$.

\section{Results and Discussion}

\subsection{Dataset}
In order to affirm the efficacy of our approach, we carried out a range of comparative experiments using the ISLES2015 SSIS MRI sequence dataset. Table 1 details the dataset, which consists of uncompressed Siemens Neuroimaging Informatics Technology Initiative (NIfTI) format files. These files share the following acquisition characteristics:
\begin{align}
    Dimensions&: 230 \times 230 \times 153 \ voxels\\
    Voxel \ size&: 1.0 mm \times 1.0 mm \times 1.0 mm\\
    Slice \ spacing&: 1.0 mm
\end{align}
The acute ischemic lesions were meticulously annotated by at least three specialists. We used 21 cases (encompassing 3213 images) for training and validating the model, while the remaining 7 cases (1071 images) were reserved exclusively for testing.

\subsection{Evaluation Metrics}
We employ several criteria to assess the effectiveness of our method. The Dice coefficient serves as a metric to assess the correspondence with segmentation results provided by neurologists. It quantifies the similarity between our segmentation (X) and the ground truth (Y) by measuring their overlap and is defined as -
\begin{align}
    DICE \ (X, Y) &= \frac{2|X \cap Y|}{|X|+|Y|}\\
\end{align}
IOU is similar to Dice Coefficient, which is defined as -
\begin{align}
    IOU \ (X, Y) &= \frac{X \cap Y}{X \cup Y}\\
\end{align}
Some other important metrics that we consider are -
\begin{align}
    Accuracy &= \frac{TP + TN}{TP+FP+TN+FN}\\
    Precision &= \frac{TP}{TP+FP}\\
    Specificity &= \frac{TN}{TN+FN}\\
    Recall &= \frac{TP}{TP+FN}
\end{align}
where, TP = True Positive , FP = False Positive, FN = False Negative, and TN = True Negative

\subsection{Hyperparameters}
For the training and validation , we used an Intel processor with $128 GB$ RAM and GPU as Tesla T4. The dataset was divided into Training and validation subsets , in a ratio of $80:20$. The ADAM optimizer was utilized to train the proposed model for $200$ epochs with a batch size of $4$. The dataset was divided into separate subsets for training and validation purposes. The learning rate for training our model was set at $1e-4$.

\subsection{Experimental Results}
\begin{table}[h]
    \centering
    \begin{tabular}{|l|c|c|c|c|c|}
        \hline
        Model & Weights & Dice & Accuracy & Precision & Recall\\
        \hline
        3D UNet & N/A & 0.73\% & 49.89\% & 31.6\% & 0.37\%\\
        \hline
        ResNet50 (X-axis) & Present & 80.7\% & 52.53\% & 61.7\% & 44.6\%\\
        \hline
        ResNet50 (Y-axis) & Present & 80.12\% & 57.89\% & 74.86\% & 65.29\%\\
        \hline
        ResNet50 (Z-axis) & Present & 21.064\% & 49.89\% & 32.6\% & 15.58\%\\
         & Absent & 43.26\% & 52.67\% & 46.57\% & 45.83\%\\
        \hline
        Proposed Model & N/A & 80.5\% & 74.03\% & 73.33\% & 62.47\%\\
        \hline
    \end{tabular}
\end{table}
This table shows the improved performance compared to the DSN method where the Dice Score was 43.26\% and accuracy was 52.67\% and our model’s provided improved Dice Score of 80.5\% and accuracy of 74.03\%.
\begin{figure}[h]
    \centering
    \includegraphics[width=0.425\textwidth]{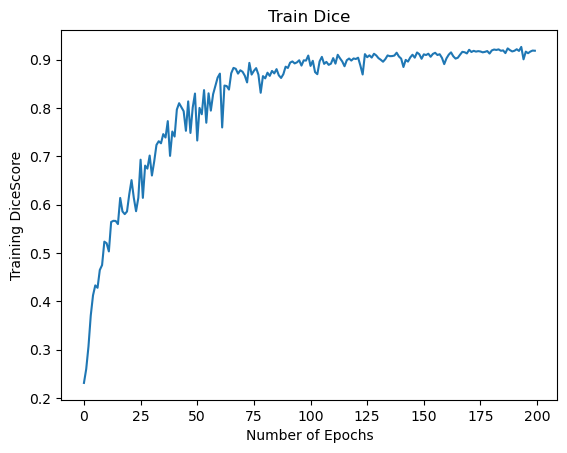}
\end{figure}\\
\begin{figure}[h]
    \centering
    \includegraphics[width=0.425\textwidth]{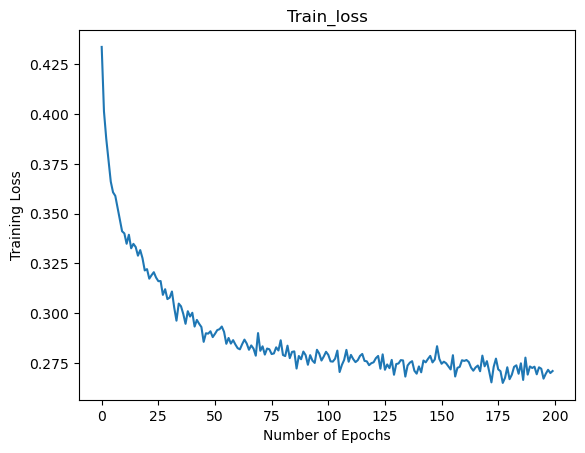}
    \includegraphics[width=0.425\textwidth]{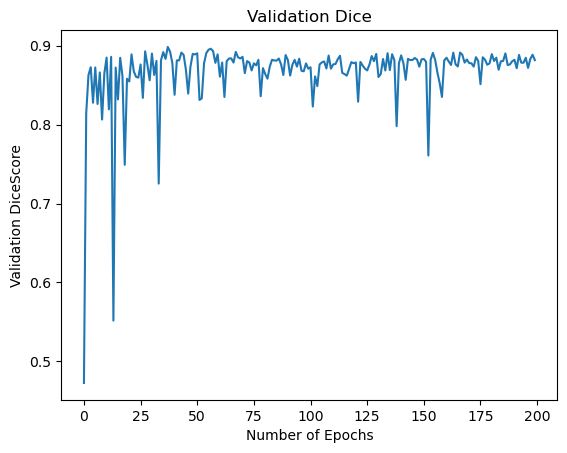}
    \includegraphics[width=0.425\textwidth]{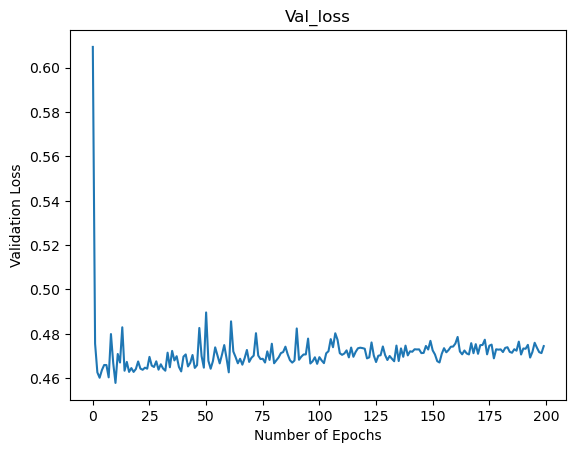}
    \caption{Graphs of Dice Score and Loss while training and validation of our proposed model after applying the Majority Voting algorithm}
\end{figure}
\begin{figure}[h]
    \centering
    \includegraphics[width=0.5\textwidth]{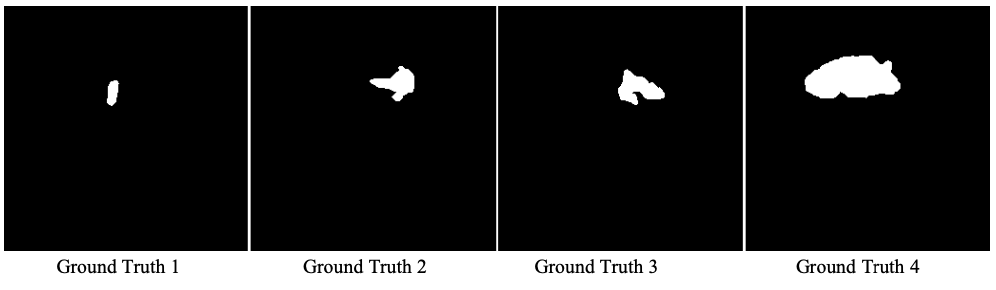}
    \includegraphics[width=0.5\textwidth]{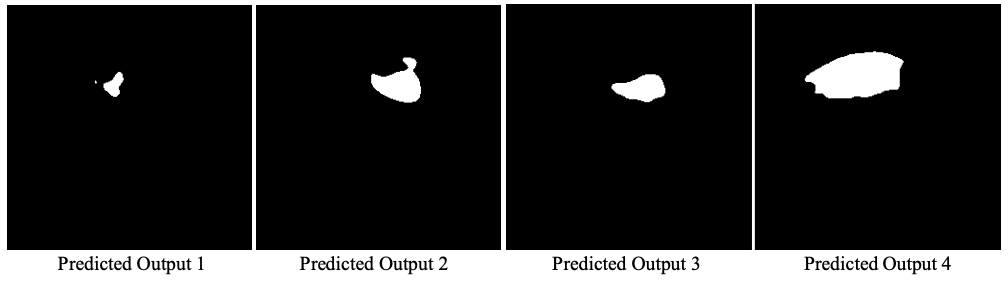}
    \includegraphics[width=0.5\textwidth]{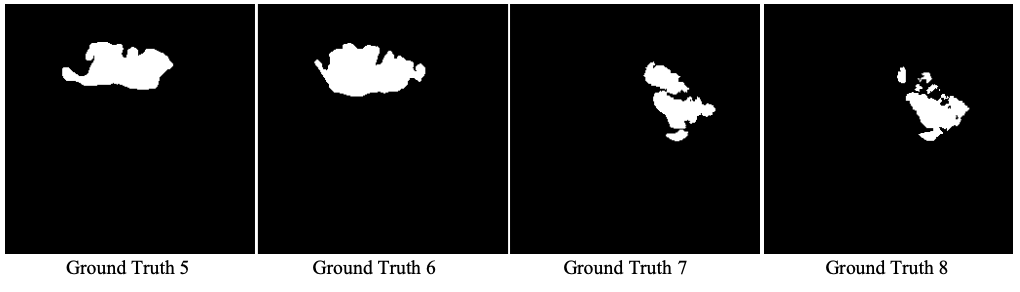}
    \includegraphics[width=0.5\textwidth]{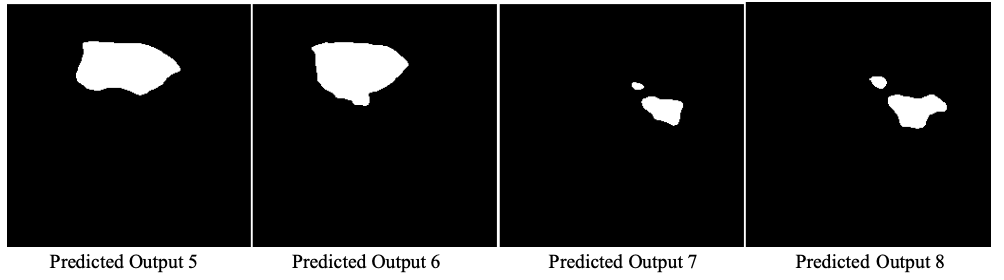}
    \caption{Some of the Predicted outputs and their corresponding Ground truths. (Note the white patches which are the lesions in the Brain. The Ground Truth is the actual Lesion image whereas the White patches in the Predicted output is the predicted image of the model)}
\end{figure}

\newpage
\section{Conclusion}
In this work, we successfully developed and validated a novel deep learning framework for the image segmentation of acute ischemic stroke lesions from multi-sequence MRI. Since models based on 3D convolution have high computational cost and data demands, we opted for a computationally efficient multi-axial 2D approach. This model was trained on axial, sagittal, and coronal planes that allowed for lesion identification from three perspectives. Subsequently, a fusion network using a Majority Voting Classifier integrated the results that produced consistent and robust 3D image segmentation outcomes. We utilized the Res-UNet and Dense-UNet architectures and evaluated the benefits of transfer learning on the ISLES2015 dataset. This automated and relatively resource-light framework addressed crucial challenges that reduced the need for manual segmentation while providing the required speed and precision within the narrow treatment window for stroke patients. Our experimental results showed the performance of this framework, achieving a competitive Dice score of 80.5\% and an accuracy of 74.03\%.

Despite these encouraging outcomes, there is a need for further improvements in these models. Though the multi-axial fusion approach is cost-effective, it might not be able to capture all long-range volumetric data that a 3D model would. Further development can focus on enhancing the fusion method by testing a Weighted Majority Voting Classifier to selectively prioritize predictions based on plane confidence. Also, we plan to explore using attention-based mechanisms directly in our Res-UNet architecture that could help our model to better distinguish between complex lesion boundaries and imaging artifacts. Finally, there is a need for validation of the framework's performance on a larger, multi-varied clinical dataset to verify real-world generalizability and impact.

\section*{Acknowledgment}
The authors duly acknowledge the use of computational resources provided by labs at the Electronics and Communication Department, National Institute of Technology Silchar, India. We also extend our sincere thanks to the other lab members for their valuable technical assistance with the code implementation and results generation for this study. Finally, we thank the organizers of the ISLES 2015 challenge for making the multi-sequence MRI dataset publicly available, which was essential for the development and validation of this research.

\section*{Conflict of Interest}
All the authors declare that they have no known conflicts of interest in terms of competing financial interests or personal relationships that could have an influence or are relevant to the work reported in this paper.

\end{document}